\documentclass[10pt,twocolumn,letterpaper]{article}

\usepackage{cvpr}              

\usepackage[dvipsnames]{xcolor}

\definecolor{cvprblue}{rgb}{0.21,0.49,0.74}
\usepackage[pagebackref,breaklinks,colorlinks,citecolor=cvprblue]{hyperref}

\usepackage[utf8]{inputenc} 
\usepackage[T1]{fontenc}    
\usepackage{url}            
\usepackage{booktabs}       
\usepackage{amsfonts}       
\usepackage{nicefrac}       
\usepackage{microtype}      
\usepackage{xcolor}         

\usepackage{graphicx}
\usepackage{amsmath}
\usepackage{amssymb}
\usepackage{booktabs}
\usepackage{bbm}
\usepackage{multirow}
\usepackage{textcomp}

\usepackage[capitalize]{cleveref}
\crefname{section}{Sec.}{Secs.}
\Crefname{section}{Section}{Sections}
\Crefname{table}{Table}{Tables}
\crefname{table}{Tab.}{Tabs.}

\def\paperID{9309} 
\def\confName{CVPR}
\def\confYear{2025}

\newcommand*{\affaddr}[1]{#1}

\newcommand{\name}{Video Depth Anything}

\begin{document}

\title{Video Depth Anything: Consistent Depth Estimation for Super-Long Videos}

\author{
Sili Chen \quad 
Hengkai Guo$^{\dagger}$  \quad 
Shengnan Zhu \quad 
Feihu Zhang \quad \\
Zilong Huang \quad 
Jiashi Feng \quad 
Bingyi Kang$^{\dagger}$ \vspace{2mm} \\
\affaddr{ByteDance} \\
\small{\href{https://videodepthanything.github.io/}{\ttfamily videodepthanything.github.io}}
}



\twocolumn[\maketitle\vspace{0em}
\begin{center}
    \vspace{-5mm}
    \includegraphics[width=\linewidth]{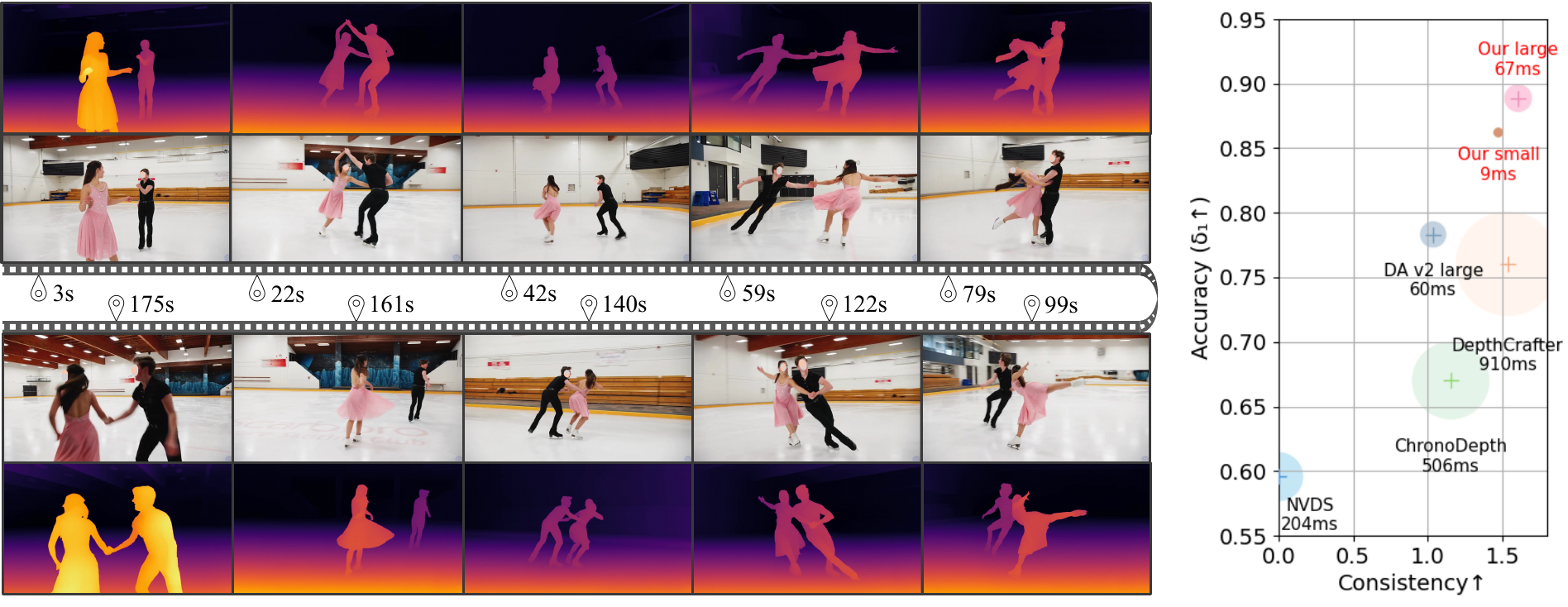}
    \vspace{-7mm}
\end{center}
\captionof{figure}{Left: Our model can generate consistent depth predictions for long videos with rich actions. The demo video shows a 196-second (4690 frames) long take of pair skating, as sourced from~\cite{huang2024wildavatar}. Right: Comparison to baselines in terms of accuracy ($\delta_1$), consistency, and latency on the Nvidia A100 GPU (denoted with circle size). Consistency is defined as the maximum \textbf{T}emporal \textbf{A}lignment \textbf{E}rror (TAE) among all models minus the TAE of each individual model. Our model achieves the best performance in all aspects.}

\vspace{-2mm}
\label{fig:teaser}\bigbreak]

\footnotetext{~$^\dagger$Corresponding author.}

\begin{abstract}
   Depth Anything has achieved remarkable success in monocular depth estimation with strong generalization ability. However, it suffers from temporal inconsistency in videos, hindering its practical applications. Various methods have been proposed to alleviate this issue by leveraging video generation models or introducing priors from optical flow and camera poses. Nonetheless, these methods are only applicable to short videos ($\textless$ 10 seconds) and require a trade-off between quality and computational efficiency. We propose Video Depth Anything for high-quality, consistent depth estimation in super-long videos (over several minutes) without sacrificing efficiency. We base our model on Depth Anything V2 and replace its head with an efficient spatial-temporal head. We design a straightforward yet effective temporal consistency loss by constraining the temporal depth gradient, eliminating the need for additional geometric priors. The model is trained on a joint dataset of video depth and unlabeled images, similar to Depth Anything V2. Moreover, a novel key-frame-based strategy is developed for long video inference.  Experiments show that our model can be applied to arbitrarily long videos without compromising quality, consistency, or generalization ability. Comprehensive evaluations on multiple video benchmarks demonstrate that our approach sets a new state-of-the-art in zero-shot video depth estimation. We offer models of different scales to support a range of scenarios, with our smallest model capable of real-time performance at 30 FPS.
\end{abstract}

\section{Introduction}
\label{sec:intro}

Recently, monocular depth estimation (MDE) has made significant progress, as evidenced by advances in depth foundation models~\cite{depth_anything_v1,depth_anything_v2,marigold,birkl2023midas}. For example, Depth Anything V2~\cite{depth_anything_v2} demonstrates a strong generalization ability in producing depth predictions with rich details in various scenarios while being computationally efficient. However, these models have a major limitation: they are mainly designed for static images and suffer from flickering and motion blur in videos. This limitation restricts their applications in robotics~\cite{dong2022towards}, augmented reality~\cite{holynski2018fast}, and advanced video editing~\cite{zhang2023controlvideo,peng2024controlnext}, which requires temporally consistent depth.

Extensive efforts are being made to address this problem. Early work~\cite{kopf2021robust,luo2020consistent,zhang2021consistent} often relies on test-time optimization to tune a pretrained monocular depth model with sophisticated geometry constraints. Given the heavy overhead at inference time of these methods, recent work mainly focuses on feedforward models and can be categorized into two approaches. The first approach~\cite{wang2023neural} involves designing a plug-and-play module to augment monocular depth model predictions with temporal consistency. The training of such a module is highly dependent on optical flow~\cite{xu2022gmflow} or camera poses~\cite{schonberger2016structure} for consistency constraints, making the module susceptible to corresponding errors. The second approach~\cite{chronodepth,hu2024depthcrafter,depthanyvideo} repurposes pre-trained video diffusion models~\cite{blattmann2023stable} into video-to-depth models. These methods excel at producing fine-grained details, but are computationally inefficient, cannot leverage existing depth foundation models, and can only handle videos with limited length. 

Then, a natural question arises: \emph{Is it possible to have a model that can perfectly inherit the capabilities of existing foundation models while achieving temporal stability for arbitrarily long videos}? In this paper, we show that the answer is YES by developing \textbf{\name} based on Depth Anything V2, without sacrificing its generalization ability, richness in details, or computational efficiency. This target is achieved without introducing any geometric priors or video generation priors.

Specifically, we first design a lightweight spatial-temporal head (STH) to replace the DPT head~\cite{ranftl2021vision} and enable temporal information interactions.
STH includes four temporal attention layers, applied along the temporal dimension for each spatial position. Introducing temporal attention only in the head prevents the learned representation from being corrupted by the limited video data. Then, we propose a \textit{temporal gradient matching loss} to constrain depth prediction gradients along the temporal dimension to match those calculated from the ground truth. This loss function is jointly optimized with the scale-shift-invariant loss and spatial gradient matching loss~\cite{birkl2023midas,depth_anything_v1}. Despite its simplicity, it can effectively boost the model's temporal consistency. Third, to maintain the original capabilities of the model, we train it jointly on 730K video frames with depth annotations using supervised learning, and on 0.62 million unlabeled images using self-training, similar to Depth Anything V2~\cite{depth_anything_v2}. To handle super-long videos at inference time, we developed a novel segment-wise processing strategy. Each new segment is concatenated with eight overlapping frames and two key frames from the previous video clips, forming a total of 32 frames. Then, the overlapping frames will be progressively interpolated between the two consecutive windows to ensure smoothness. 

We compare our model with baselines on five datasets for zero-shot video depth estimation. Our model achieves state-of-the-art (SOTA) results on four of the datasets in terms of spatial accuracy and outperforms all baselines on all datasets in terms of temporal consistency. Not only can our model produce depth outputs visually comparable to video-diffusion-based methods, but it is also significantly more computationally efficient. For the first time, we can estimate consistent depth for videos over several minutes (see \cref{fig:teaser}). Additionally, we tested our model for zero-shot image depth estimation on five datasets, noting only a marginal performance drop on one dataset compared to Depth Anything V2. As shown in \cref{fig:teaser} (right), our model achieves the best performance in all three aspects: spatial accuracy, temporal consistency, and computational efficiency.

Our contributions are summarized as follows:
\begin{itemize} 
\item We develop a novel method to transform Depth Anything into \name for depth estimation in arbitrarily long videos.
\item We propose a simple yet effective loss function that enforces temporal consistency constraints without introducing geometric or generative priors.
\item Our model not only sets new SOTA (both spatially and temporally) in video depth estimation but is also the most computationally efficient.
\end{itemize} 
\section{Related Work}
\label{sec:related_work}
{\bf Monocular depth estimation.} Early monocular depth estimation ~\cite{eigen2014depth,fu2018dorn,bhat2021adabins,yuan2022crf} efforts were primarily trained and tested on in-domain datasets. These models were constrained by the limited diversity of their training data, showing bad generalization for zero-shot application. Subsequently, MiDaS~\cite{birkl2023midas} introduced multi-dataset mixed training using an affine-invariant alignment method, significantly enhancing model generalization. However, due to limitations in the backbone model's performance and noise in the labeled depth data, the resulting depth maps lacked fine details. Following MiDaS~\cite{birkl2023midas}, monocular depth estimation models have generally bifurcated into two categories: relative depth models that estimate affine-invariant depth (e.g., DPT~\cite{ranftl2021vision}, DepthAnything~\cite{depth_anything_v1,depth_anything_v2}, Marigold~\cite{marigold}) and metric depth models that estimate depth with an absolute scale (e.g., ZoeDepth~\cite{bhat2023zoedepth}, Metric3D~\cite{yin2023metric3d}, UniDepth~\cite{piccinelli2024unidepth}). Metric depth models require training with metric depth data that includes camera parameters~\cite{yin2023metric3d}, thus their available training data is more limited compared to affine-invariant depth models, resulting in poorer generalization. Recently, Depth Anything V2~\cite{depth_anything_v2} leveraged the Dinov2~\cite{oquab2023dinov2} pre-trained backbone to train an affine-invariant large-scale model using synthetic data with high-detail fidelity. This large model was then used as a teacher to distill smaller models on 62 million unlabeled datasets~\cite{depth_anything_v1}, achieving SOTA performance in both generalization and geometric details. However, since Depth Anything V2~\cite{depth_anything_v2} was trained exclusively on static images, thus lacks temporal consistency.

{\bf Consistent video depth estimation.} The core task of consistent video depth estimation is to obtain temporal consistent and accuracy depth maps. Early methods for video depth relied on test-time training~\cite{luo2020consistent,kopf2021robust,zhang2021consistent}, which were impractical for applications for their low efficiency. In recent years, learning-based models have emerged. Some of these models, such as MAMo~\cite{yasarla2023mamo}, use optical flow~\cite{teed2020raft} to warp features, while others~\cite{sayed2022simplerecon} depends on relative poses between frames to construct cost volumes. However, their performance were suffered from errors of inaccurate optical flow or pose estimation. Additional approaches have attempted to enhance off-the-shelf monocular depth estimation (MDE) models with temporal stability model blocks ~\cite{wang2023neural}. Nevertheless, these efforts have not achieved satisfactory results due to suboptimal model designs and inadequate geometric consistency constraints. Furthermore, video diffusion models such as ChronoDepth~\cite{chronodepth}, DepthCrafter~\cite{hu2024depthcrafter}, and DepthAnyVideo\cite{depthanyvideo} show better details and temporal consistency. But they suffered from slow inference speeds and require extensive video depth training data. Limited by the large memory, these models~\cite{hu2024depthcrafter} were typically tested only within the maximum window length used during training, leading to depth flickering between windows and poor temporal and spatial consistency in long videos.
\section{Video Depth Anything}
\label{sec:method}

\begin{figure*}[t]
  \centering
   \includegraphics[width=1.0\linewidth]{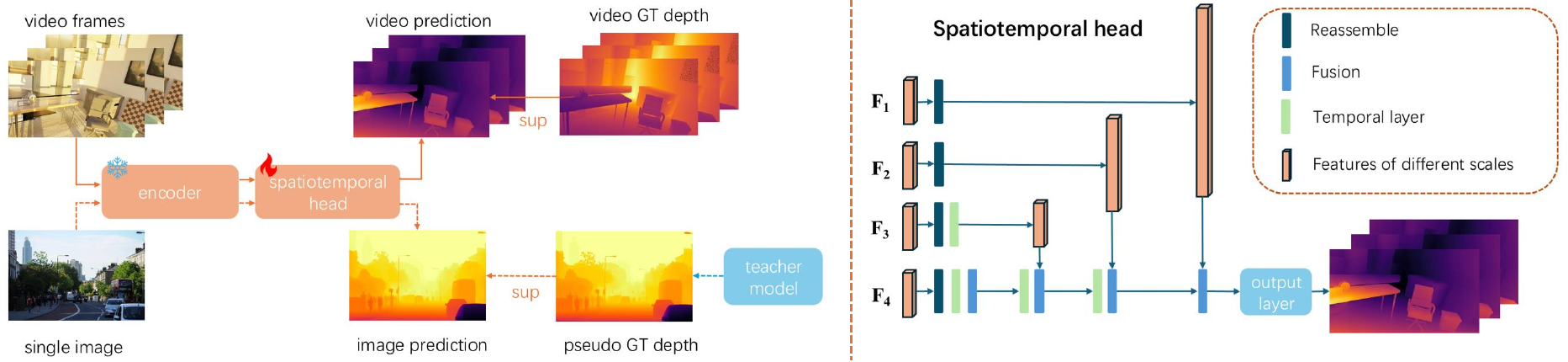}

   \caption{\textbf{Overall pipeline and the spatio-temporal head}. Left: Our model is composed of a backbone encoder from Depth Anything V2 and a newly proposed spatio-temporal head. We jointly train our model on video data using ground-truth depth labels for supervision and on unlabeled images with pseudo labels generated by a teacher model. During training, only the head is learned. Right: Our spatiotemporal head inserts several temporal layers into the DPT head, while preserving the original structure of DPT head~\cite{ranftl2021vision}. 
   }
   \label{fig:overview_head}
   \vspace{-18pt}
\end{figure*}

In this section, we introduce Video Depth Anything, a feed-forward video transformer model to efficiently estimate temporally consistent video depth. We adopt the affine-invariant depth, but share the same scale and shift across the entire video. The pipeline of our method is shown in Fig.~\ref{fig:overview_head}. Our model is built upon Depth Anything V2 with an additional temporal module and video dataset training (Sec.~\ref{subsec::video}). A novel loss to enfoce temporal consistency is proposed in Sec.~\ref{subsec::loss}. Finally, a strategy combined with overlapping frames and key frames is presented to efficiently support super-long video inference (Sect.~\ref{subsec::long}). 

\subsection{Architecture}
\label{subsec::video}

Due to the lack of sufficient video depth data, we start with a pre-trained image depth estimation model, Depth Anything V2, and adopt a joint training strategy using both image and video data.

\noindent{\bf Depth Anything V2 Encoder.} Depth Anything V2~\cite{depth_anything_v2} is the current state-of-the-art monocular depth estimation model, characterized by its high accuracy and generalization capabilities. We use its trained model as our encoder. To reduce training costs and preserve well-learned features, the encoder is frozen during training.

Unlike monocular depth encoders that only accept image input, our training scenario requires the encoder to process simultaneously both video and image data. To extract features from video frames with an image encoder, we collapse the temporal dimension of a video clip into the batch dimension. The input data are denoted as $\mathbf{X} \in \mathbb{R}^{(B \times N) \times C \times H \times W} $, where $B$ represents the batch size, $N$ is the number of frames in the video clip, $N=1$ for the image as input, $C,H,W$ are the number of channels, height, width of the frames, respectively. The encoder takes $\mathbf{X}$ as input to produce a series of intermediate feature maps $\mathbf{F_i} \in \mathbb{R}^{(B \times N) \times (\frac{H}{p} \times \frac{W}{p}) \times C_i }$, $p$ is the patch size of the encoder. Although the image encoder extracts strong visual representations from individual frames, it neglects the temporal information interactions between frames. Thus, the spatiotemporal head is introduced to model the temporal relationship among the frames.

\noindent{\bf Spatiotemporal Head.}
The spatiotemporal head (STH) is built upon the DPT~\cite{ranftl2021vision} head and with the only modification being the insertion of temporal layers to capture temporal information.
A temporal layer consists of a multi-head self-attention~\cite{vaswani2017attention} model (SA) and a feed-forward network (FFN). When inputting a feature $\mathbf{F_i}$ into the temporal layer, the temporal dimension $N$ is isolated, and self-attention is executed solely along the temporal dimension to facilitate the interaction of temporal features. To capture temporal positional relationships among different frames, we utilize absolute positional embedding to encode temporal positional information from the video sequence.

The spatiotemporal head uniformly samples 4 feature maps from $\mathbf{F_i}$ (including the final features from the encoder, denoted as $\mathbf{F_{4}}$) as inputs, and predicts a depth map $\mathbf{D} \in \mathbb{R}^{H \times W}$. As shown in Figure~\ref{fig:overview_head}, the selected features $\mathbf{F_i}$ are fed into the Reassemble layer to produce a feature pyramid. Then, the features are gradually fused from low resolution to high resolution by the Fusion layer. The Reassemble and Fusion layer are proposed by DPT~\cite{ranftl2021vision}. The final fused high-resolution feature maps are passed through the output layer to produce the depth map $\mathbf{D}$. To reduce the additional computational load, we insert the temporal layer at a few positions with lower feature resolutions. 

\subsection{Temporal Gradient Matching loss}
\label{subsec::loss}

In this section, we start with the Optical Flow Based Warping (OPW) loss, then explore new loss designs and ultimately propose a Temporal Gradient Matching Loss (TGM) that does not rely on optical flow, yet still ensures the temporal consistency of predictions between frames.

\noindent{\bf OPW loss.} To constrain temporal consistency, previous video models such as~\cite{wang2023neural,10.1145/3591106.3592264,Wang_2022} assume that the depths at corresponding positions in adjacent frames, identified through optical flow, are consistent, \textit{e.g.}, the Optical Flow based Warping (OPW) loss proposed in NVDS~\cite{wang2023neural}. OPW loss is computed after obtaining corresponding points on the basis of optical flow and warping. Specifically, for two consecutive depth prediction results, $p_i$ and $p_{i+1}$. $p_{i+1}$ is warped to $\hat{p_i}$ according to the wrapping relationship derived from the optical flow, and then the loss is calculated with:
\vspace{-3mm}
\begin{equation}
  \mathcal{L}_\text{OPW}= \frac {1}{N-1} \sum _ {i=2}^ {N} \parallel p_i - \hat{p_i}\parallel_1,
  \label{eq:opw_loss}
\vspace{-3mm} 
\end{equation}
where $N$ denotes the length of a video window, and $||\cdot||_1$ represents $\ell$1 distance. However, there is a fundamental issue with the OPW loss: the depth of corresponding points is not invariant across adjacent frames. This assumption holds true only when adjacent frames are stationary. For instance, in driving scenario, when a car is moving forward, the distance to static objects in front decreases relative to the car, violating the assumption of $\mathcal{L}_{OPW}$. To address this inherent issue of OPW, we propose a new loss function to constrain the temporal consistency of depth.

\noindent{\bf Temporal gradient matching loss (TGM)}. When calculating the loss, we do not assume that the depth of the corresponding points in adjacent frames remains unchanged. Instead, we posit that the change in depth of corresponding points between adjacent prediction frames should be consistent with the change observed in ground truth. We refer to this discrepancy as a stable error (SE) given by: 
\vspace{-3mm}
\begin{equation}
    \mathcal{L}_\text{SE} = \frac{1}{N-1} \sum_{i=1}^{N-1}\parallel \mid\hat{d_i}-d_{i}\mid - \mid\hat{g_i}-g_{i}\mid \parallel_1.
  \label{eq:stable_loss}
  \vspace{-3mm}
\end{equation}
Here, $d_i,g_i$ are scaled and shifted versions of the predictions and ground truth. $\hat{d_i},\hat{g_i}$ denotes the warped depth from the subsequent frame using optical flow. $\mid\cdot\mid$ is used to represent the absolute values. 

However, generating optical flow incurs additional overhead. To address the dependence on optical flow, we further generalize the above assumption. Specifically, it is not necessary to use the corresponding points obtained from the optical flow. Instead, we directly use the depth at the same coordinate in adjacent frames to calculate the loss. The assumption is that the change in depth at the same image position between adjacent frames should be consistent with that in the ground truth. Since this process is akin to calculating the gradient of values in temporal dimension, we name it Temporal Gradient Matching Loss, as given by
\vspace{-3mm}
\begin{equation}
    \mathcal{L}_\text{TGM} = \frac{1}{N-1} \sum_{i=1}^{N-1}\parallel \mid d_{i+1}-d_{i}\mid - \mid g_{i+1}-g_{i}\mid \parallel_1.
    \vspace{-3mm}
\end{equation}
In practice, we only compute the TGM loss in regions where the change in ground truth depth, \textit{i.e.}, $\mid g_{i + 1} - g_i \mid < 0.05 $. This threshold helps to avoid sudden changes in depth map caused by edges, dynamic objects, and other factors that introduce unsteadiness during training. 

Our total loss to supervise video depth data is as follows:
\vspace{-3mm}
\begin{equation}
    \mathcal{L}_{\text{all}} = \alpha \mathcal{L}_{\text{TGM}} + \beta  \mathcal{L}_{\text{ssi}},
  \label{eq:all_loss}
  \vspace{-3mm}
\end{equation}
where $\mathcal{L}_{ssi}$ is a scale- and shift-invariant loss to supervise single images proposed by MiDaS~\cite{birkl2023midas}. $\alpha$ and $\beta$ are weights to balance spatio-temporal consistency and spatial structure in a single frame. 

\subsection{Inference strategy for super-long sequence}
\label{subsec::long}

\begin{figure}
    \centering
    \includegraphics[width=\linewidth]{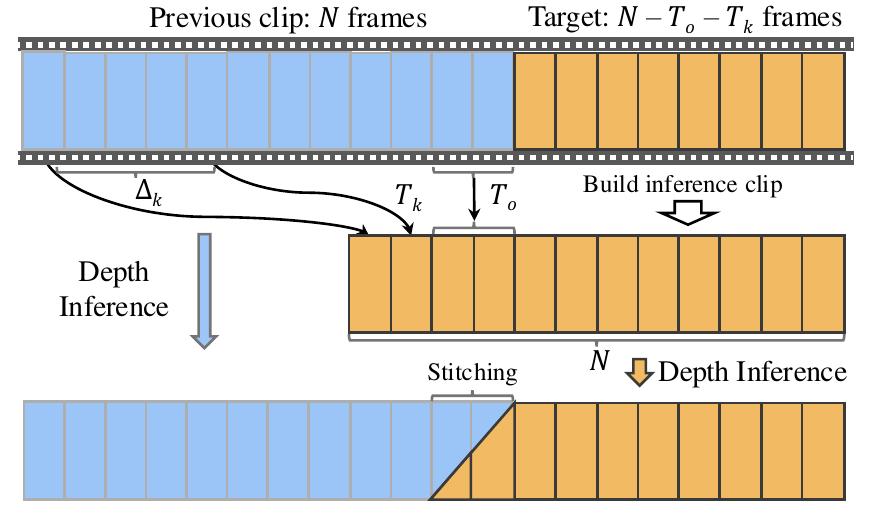}
    \caption{\textbf{Inference strategy for long videos}. $N$ is the video clip lenght consumed by our model. Each inference video clip is built by $N-T_o-T_k$ future frames, $T_o$ overlapping/adjacent frames, and $T_k$ key frames. The key frames are selected by taking every $\Delta_k$-th frame going backward. Then, the new depth predictions will be scale-shift-aligned to the previous frames based on the $T_k$ overlapping frames. We use $N=32, T_o=8, T_k=2, \Delta_k=12$.}
    \label{fig:long_inference}
    \vspace{-18pt}
\end{figure}
To handle videos of arbitrary length, a straightforward approach is simply to concatenate the model outputs from different video windows. However, this method fails to ensure smooth transitions between windows. A more sophisticated technique entails inferring video windows with overlapping regions. By utilizing the predicted depth of the overlapping regions to compute an affine transform, predictions from one window can be aligned with those from another. Nevertheless, this method can introduce accumulated errors through successive affine alignments, leading to depth drift in extended videos. To address these challenges in ultra-long videos with a limited inference window size, we proposed key-frame referencing to inherit scale and shift information from past predictions and overlapping interpolation to ensure smooth inference across local windows.

\noindent{\bf Key-frame referencing.} As illustrated in ~\cref{fig:long_inference}, a subsequent video clip for inference is composed of three parts: $N-T_o-T_k$ future frames, $T_o$ overlapping frames from the previous clip and $T_k$ \textbf{key frames}. The key frames are subsampled from the previous frames with an interval of size $\Delta_k$. Therefore, the video clip to be consumed share the same length as during training. This approach incorporates content from earlier windows into the current window with minimal computation burden. Furthermore, we carefully select the values of $T_k$ and $\Delta_k$ to ensure that the first frame of a video is always positioned at the beginning of each clip, thereby enhancing depth consistency for extended videos. According to our experiment results, such simple strategy can significantly reduce accumulated scale drift, especially for long video.

\noindent{\bf Depth clip stitching.} Using $T_o$ overlapping frames (in ~\cref{fig:long_inference}) between two consecutive windows is crucial for avoiding flickering depth predictions. 
The effects of overlapping frames are twofold. 
First, by sharing partial frame features, the scale and shift across consecutive windows will be more similar. Second, the depth prediction for the overlapping frames is updated by interpolating between the two segments. Assume the depth for the $o_i$-th overlapping frame from the previous segment is denoted by $\mathbf{D}_{o_i}^{\text{pre}}$, and the depth from the current segment is denoted by $\mathbf{D}_{o_i}^{\text{cur}}$. The final depth is updated as $\mathbf{D}_{o_i} = \mathbf{D}_{o_i}^{\text{pre}} \cdot w_i + \mathbf{D}_{o_i}^{\text{cur}} \cdot (1 - w_i)$, where $w_i$ linearly decays from 1 to 0 as $i$ increases from 1 to $T_o$.
\section{Experiments}
\label{sec:experiments}
\begin{table*}[]
\centering
\resizebox{0.9\linewidth}{!}{
\begin{tabular}{lcccccccccccc}
\toprule
\centering
\multirow{2}{*}{Method / Metrics} & \multicolumn{2}{c}{KITTI~\cite{geiger2013vision}} & \multicolumn{2}{c}{Scannet~\cite{dai2017scannet}} & \multicolumn{2}{c}{Bonn~\cite{palazzolo2019iros}} & \multicolumn{2}{c}{NYUv2~\cite{Silberman:ECCV12}} & \multicolumn{2}{c}{Sintel~\cite{Butler_Wulff_Stanley_Black_2012}(\raisebox{0.5ex}{\texttildelow}50 frames)} & \multicolumn{1}{c}{Scannet (170 frames\cite{depthanyvideo})} & \multirow{2}{*}{$\delta_1$ Rank} \\ 
\cmidrule(r){2-12} 
& AbsRel~(↓) & $\delta_1$~(↑) & AbsRel~(↓) & $\delta_1$~~(↑) & AbsRel~(↓) & $\delta_1$~~(↑) & AbsRel~(↓) & $\delta_1$~~(↑) & AbsRel~(↓) & $\delta_1$~~(↑) & TAE~(↓) \\
\midrule 
DAv2-L~\cite{depth_anything_v2} & 0.137  & 0.815  & 0.150  & 0.768  & 0.127  & 0.864  & 0.094  & 0.928  & 0.390 & 0.541 & 1.140 & 5.5 \\
NVDS~\cite{wang2023neural} & 0.233  & 0.614  & 0.207  & 0.628  & 0.199  & 0.674  & 0.217  & 0.598  & 0.408  & 0.464  & 
 2.176 & 8.5 \\
NVDS~\cite{wang2023neural} + DAv2-L~\cite{depth_anything_v2} & 0.227  & 0.617  & 0.194  & 0.658  & 0.191  & 0.700  & 0.184  & 0.679  & 0.449  & 0.503  & 2.536 & 7.7 \\
ChoronDepth~\cite{chronodepth} & 0.243 & 0.576 & 0.190 & 0.682 & 0.159 & 0.793 & 0.199 & 0.665 & 0.551 & 0.478 & 1.022 & 6.6 \\
DepthCrafter~\cite{hu2024depthcrafter} & 0.164  & 0.753  & 0.169  & 0.730  & 0.153  & 0.803  & 0.141  & 0.822  & \underline{0.299}  & \textbf{0.695}  & 0.639 & 5.0 \\
DepthAnyVideo~\cite{depthanyvideo} & -  & -  & -  & -  & -  & -  & -  & -  & 0.405  & \underline{0.659}  & 0.967 & - \\
VDA-S (Ours-Syn) & 0.089          & 0.937             & 0.124             & 0.839             & 0.081             & 0.955             & 0.087             & 0.953             & 0.326             & 0.596 & 0.702          & 4.0 \\
VDA-L (Ours-Syn) & \textbf{0.083} & \textbf{0.946}    & \textbf{0.087}    & \textbf{0.933}    & \textbf{0.070}    & \textbf{0.961}    & \underline{0.064} & \underline{0.967} & 0.300             & 0.633 & \textbf{0.570} & \textbf{1.7}\\
VDA-S (Ours)     & 0.086          & 0.942             & 0.110             & 0.876             & 0.083             & 0.950             & 0.077             & 0.959             & 0.339             & 0.584 & 0.703          & 3.9 \\
VDA-L (Ours)     & \textbf{0.083} & \underline{0.944} & \underline{0.089} & \underline{0.926} & \underline{0.071} & \underline{0.959} & \textbf{0.062}    & \textbf{0.971}    & \textbf{0.295} & 0.644 & \textbf{0.570} & \textbf{1.7}\\
\bottomrule
\end{tabular}}
\caption{\textbf{Zero-shot video depth estimation results}. We compare with representative single-image~\cite{depth_anything_v2} and video depth estimation models~\cite{wang2023neural, chronodepth, hu2024depthcrafter, depthanyvideo}. ``VDA-S'' and ``VDA-L'' denote our model with ViT-Small and ViT-Large backbones, respectively. Models denoted with ``-Syn'' are trained exclusively on public datasets, as detailed in supplementary. The \textbf{best} and the \underline{second best} results are highlighted.}
\label{tab::quant_video_depth_benchmark}
\vspace{-8pt}
\end{table*}

\begin{table*}[h]
\centering
\resizebox{0.9\linewidth}{!}{
\begin{tabular}{lccccccccccc}
\toprule
\centering
\multirow{2}{*}{Method / Metrics} & \multicolumn{2}{c}{KITTI} & \multicolumn{2}{c}{Sintel} & \multicolumn{2}{c}{NYUv2} & \multicolumn{2}{c}{ETH3D} & \multicolumn{2}{c}{DIODE} & \multirow{2}{*}{$\delta_1$ Rank}\\ 
\cmidrule(r){2-11} 
& AbsRel~(↓) & $\delta_1$~~(↑) & AbsRel~(↓) & $\delta_1$~~(↑) & AbsRel~(↓) & $\delta_1$~~(↑) & AbsRel~(↓) & $\delta_1$~~(↑) & AbsRel~(↓) & $\delta_1$~~(↑) \\
\midrule 
DepthCrafter~\cite{hu2024depthcrafter} & 0.107 & 0.891 & 0.568 & 0.652 & 0.082 & 0.936 & 0.179 & 0.793 & 0.141 & 0.857 & 4 \\
DepthAnyVideo~\cite{depthanyvideo} & \textbf{0.073} & \textbf{0.946} & 0.687 & 0.692 & 0.058 & 0.963 & \textbf{0.123} & \textbf{0.881} & 0.072 & 0.942 & 2.4 \\
DAv2-L~\cite{depth_anything_v2} & \underline{0.074} & \textbf{0.946} & \textbf{0.487} & \underline{0.752} & \textbf{0.045} & \textbf{0.979} & \underline{0.131} & \underline{0.865} & \textbf{0.066} & \textbf{0.952} & \textbf{1.4} \\
VDA-L (Ours) & 0.075 & \textbf{0.946} & \underline{0.496} & \textbf{0.754} & \underline{0.046} & \underline{0.978} & 0.132 & 0.863 & \underline{0.067} & \underline{0.950} & \underline{2} \\
\bottomrule
\end{tabular}}
\caption{\textbf{Zero-shot single-image depth estimation results}. We compare with representative single-image~\cite{depth_anything_v2} and video depth estimation models~\cite{hu2024depthcrafter, depthanyvideo} with single-frame inputs. ``VDA-L'' denotes our model with ViT-Large backbone. The \textbf{best} and the \underline{second best} results are highlighted.}
\label{tab::quant_monocular_depth_benchmark}
\vspace{-18pt}
\end{table*}

\subsection{Evaluation}
\noindent{\bf Datasets.} For the quantitative evaluation of video depth estimation, we utilize five datasets that encompass a wide range of scenes, including indoor~\cite{dai2017scannet,Silberman:ECCV12,palazzolo2019iros}, outdoor~\cite{geiger2013vision}, and wild environments~\cite{Butler_Wulff_Stanley_Black_2012}. Each video is evaluated using up to 500 frames, which is significantly more extensive than the 110 frames used in ~\cite{hu2024depthcrafter}. For results with 110 frames, see the appendix for details. In addition to video depth evaluation, we also assess our model's performance on static images~\cite{depth_anything_v2} on five image benchmarks ~\cite{geiger2013vision,Butler_Wulff_Stanley_Black_2012,Silberman:ECCV12,Schops_Schonberger_Galliani_Sattler_Schindler_Pollefeys_Geiger_2017,Igor_Nicholas_Zhang_Luo_Wang_Dai_Daniele_Mostajabi_Basart_Walter_et_al_2019}.

\noindent{\bf Metrics.} We evaluate our video depth model using both geometric accuracy and temporal stability metrics. In accordance with ~\cite{hu2024depthcrafter}, we first align the predicted depth maps with the ground truth by applying a uniform scale and shift throughout the video. For geometric accuracy, we compute the Absolute Relative Error (AbsRel) and $\delta_1$ metrics ~\cite{depth_anything_v2,hu2024depthcrafter}. To assess temporal stability, we use the Temporal Alignment Error (TAE) metric in ~\cite{depthanyvideo}, to measure the reprojection error of the depth maps between consecutive frames.

\subsection{Zero-shot Depth Estimation}
We compare our model against four representative video depth estimation models: NVDS~\cite{wang2023neural}, ChronoDepth~\cite{chronodepth}, DepthCrafter~\cite{hu2024depthcrafter}, and DepthAnyVideo~\cite{depthanyvideo} on established video depth benchmarks. 
Furthermore, we introduce two robust baselines, 1) Depth Anything V2~\cite{depth_anything_v2} (DAv2), and 2) NVDS + DAv2, \textit{i.e.}, replacing the base model in NVDS with DAv2. 
It is important to note that DepthAnyVideo~\cite{depthanyvideo} supports a maximum of 192 frames per video; therefore, we report metrics for the Sintel~\cite{Butler_Wulff_Stanley_Black_2012} dataset exclusively for this model, as other datasets contain videos that exceed this frame limit. For static image evaluation, we compare the performance of our model with DepthCrafter~\cite{hu2024depthcrafter}, DepthAnyVideo~\cite{depthanyvideo}, and Depth Anything V2~\cite{depth_anything_v2}.

\noindent\textbf{Video depth results.}
As demonstrated in Table~\ref{tab::quant_video_depth_benchmark}, our VDA model achieves state-of-the-art performance across all long video datasets, excelling in both geometric and temporal metrics. This underscores the effectiveness of our robust foundation model and the innovative design of our video model. Notably, on the KITTI~\cite{geiger2013vision}, Scannet~\cite{dai2017scannet}, and Bonn~\cite{palazzolo2019iros} datasets, our model surpasses other leading methods by a significant margin of approximately $10\%$ in the geometric accuracy metric $\delta_1$, although it is trained on much fewer video data compared to DepthCrafter~\cite{hu2024depthcrafter} (over 10 million frames) and DepthAnyVideo \cite{depthanyvideo} (6 million frames). For the short video synthetic dataset Sintel~\cite{Butler_Wulff_Stanley_Black_2012}, where sequences contain around 50 frames each, DepthCrafter~\cite{hu2024depthcrafter} exhibits better accuracy than our model. This discrepancy may be attributed to the absence of movie data, which features frames with focal lengths similar to those in Sintel~\cite{Butler_Wulff_Stanley_Black_2012}, in our model's training set. It is also worth highlighting that our compact model, VDA-S, which has significantly lower latency compared to other models (as shown in Table~\ref{tab::infer_time}), demonstrates superior geometric accuracy over representative diffusion-based methods for long videos.

\noindent\textbf{Image depth results.} The datasets in Table~\ref{tab::quant_monocular_depth_benchmark} consist of single images, identical to those utilized in DAv2~\cite{depth_anything_v2}. Although our model is designed for videos, it demonstrates the capability to handle static images. As shown in Table~\ref{tab::quant_monocular_depth_benchmark}, our video depth model achieves competitive depth metrics compared to DAv2-L~\cite{depth_anything_v2} across most datasets. This demonstrates that our model not only retains the geometric accuracy of the foundational model but also delivers high accuracy and consistency in video depth estimation.

\begin{figure}[t]
  \centering
   \includegraphics[width=1.0\linewidth]{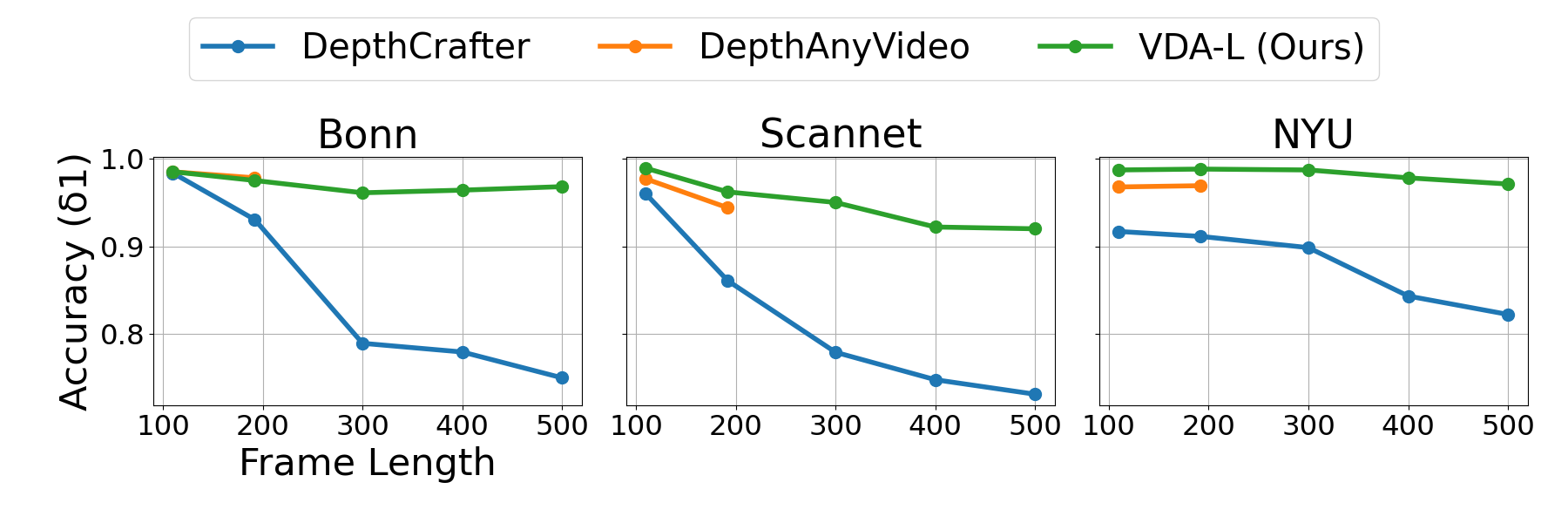}
   \vspace{-18pt}
   
   \caption{\textbf{Video depth estimation accuracy for different frame length.} We compare our model (VDA-L) with DepthCrafter~\cite{hu2024depthcrafter} and DepthAnyVideo~\cite{depthanyvideo} from 110 to 500 frames on Bonn~\cite{palazzolo2019iros}, Scannet~\cite{dai2017scannet}, and NYUv2~\cite{Silberman:ECCV12}.}
   \label{fig:long_video_decay}
   \vspace{-18pt}
\end{figure}

\noindent\textbf{Long video quantitative results.} We selected 10 scenes each from Bonn~\cite{palazzolo2019iros} and Scannet~\cite{dai2017scannet}, and 8 scenes from NYUv2~\cite{Silberman:ECCV12}, with each scene comprising 500 video frames. We then evaluated the video depth at frame lengths of 110, 192, 300, 400, and 500, where 110 and 192 correspond to the maximum window sizes of DepthCrafter~\cite{hu2024depthcrafter} and DepthAnyVideo~\cite{depthanyvideo}, respectively. The variation in metrics is shown in Figure~\ref{fig:long_video_decay}. As illustrated, our model significantly outperforms DepthCrafter~\cite{hu2024depthcrafter} in all evaluated frame lengths for all datasets, exhibiting minimal metric degradation as the number of frames increases. Furthermore, our model surpasses DepthAnyVideo~\cite{depthanyvideo} in Scannet~\cite{dai2017scannet} and NYUv2~\cite{Silberman:ECCV12}, and achieves comparable results in Bonn~\cite{palazzolo2019iros} for the 110 and 192 frame metrics. Most notably, our approach supports inference for arbitrarily long videos, providing a substantial advantage in practical applications.

\begin{figure}
    \centering
    \includegraphics[width=\linewidth]{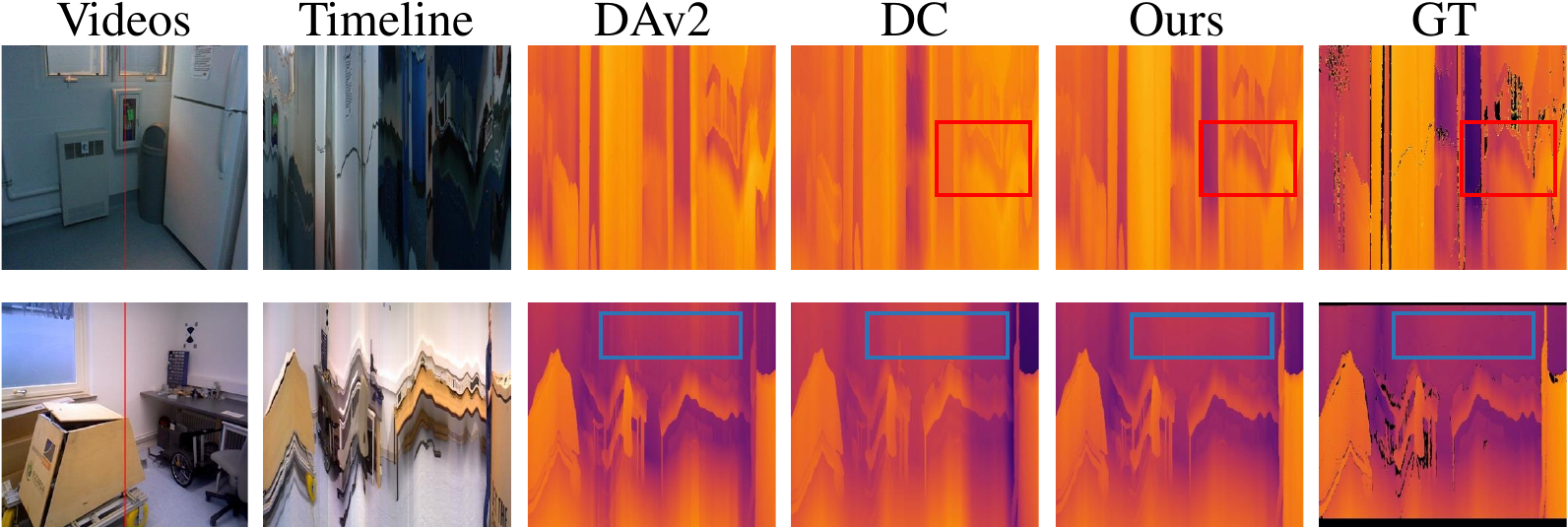}
    \caption{\textbf{Qualitative comparison for real-world long video depth estimation.} We compare our model with DAv2-L~\cite{depth_anything_v2} and DepthCrafter~\cite{hu2024depthcrafter} on 500-frame videos from Scannet~\cite{dai2017scannet} and Bonn~\cite{palazzolo2019iros}. 
    }
    \label{fig:long_video_show}
    \vspace{-18pt}
\end{figure}

\noindent\textbf{Qualitative results.} We present two results of the long video visualization in~\ref{fig:long_video_show}. The second column represents image temporal profiles obtained by slicing images along the timeline at the red-line positions. The subsequent columns represent the corresponding depth profiles. The red boxes highlight instances where the depth profile of our model more closely resembles the ground truth (GT) compared to DepthCrafter~\cite{hu2024depthcrafter}, indicating superior geometric accuracy. Furthermore, our model demonstrates better temporal consistency, as shown in the blue boxes. In these instances, DepthCrafter~\cite{hu2024depthcrafter} exhibits drifted depth, and DAv2-L~\cite{depth_anything_v2} produces flickering depth. 

\begin{figure*}[t]
  \centering
   \includegraphics[width=0.95\linewidth]{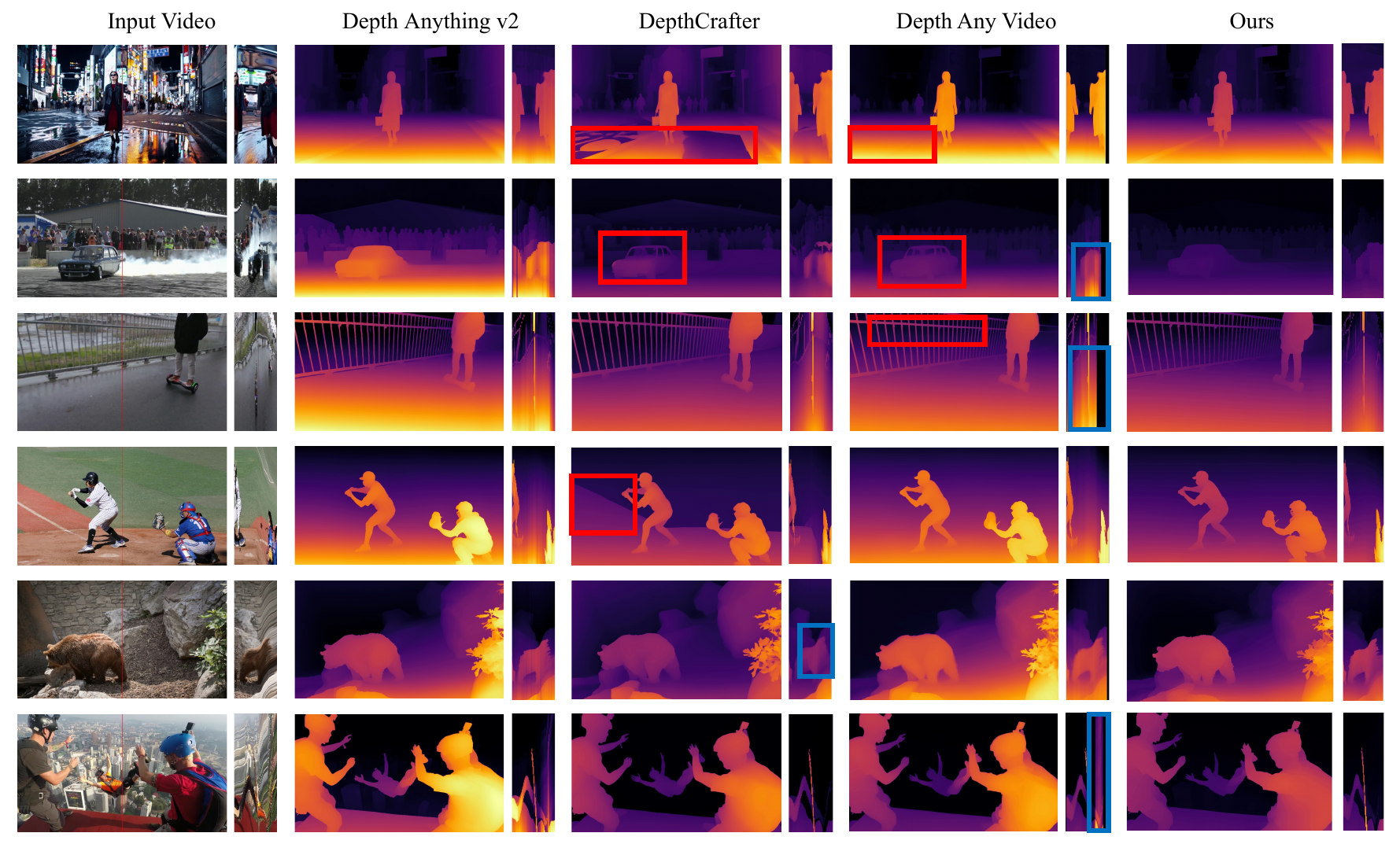}
   \vspace{-8pt}

   \caption{\textbf
   {Qualitative comparison for in-the-wild short video depth estimation.} We compare with Depth-Anything-V2~\cite{depth_anything_v2}, DepthCrafter~\cite{hu2024depthcrafter} and DepthAnyVideo~\cite{depthanyvideo} on videos with less than 100 frames from DAVIS~\cite{Perazzi2016davis}. Red boxes show incorrect depth estimation while blue boxes show inconsistent depth estimation.
   }
   \label{fig:short_video_show}
   \vspace{-18pt}
\end{figure*}

In addition to long videos, we present in-the-wild short video results in Figure~\ref{fig:short_video_show}. Depth Any Video~\cite{depthanyvideo} exhibits depth inconsistency even within a single reference window, as indicated by the blue boxes. Although DepthCrafter~\cite{hu2024depthcrafter} demonstrates smoother depth along video frames compared to Depth Any Video~\cite{depthanyvideo}, it fails to estimate accurate depth in some complex environments.

\noindent{\bf{Inference time.}} We measure the inference latency of various models on an A100 GPU. As shown in Table~\ref{tab::infer_time}, our large model achieves the lowest inference time compared to both diffusion-based methods (DepthAnyVideo~\cite{depthanyvideo} and DepthCrafter~\cite{hu2024depthcrafter}) and the transformer-based method (NVDS~\cite{wang2023neural}). This performance is attributed to our feed-forward transformer structure and lightweight temporal modules. Notably, the latency of our large model, VDA-L, is only approximately $10\%$ greater than that of DAv2-L~\cite{depth_anything_v2,hu2024depthcrafter}, which uses the same encoder structure, thus demonstrating the efficiency of our spatiotemporal head. Furthermore, the inference latency of our small model is less than 10ms, indicating its potential for real-time applications.
\begin{table}[]
\centering
\begin{tabular}{lcc}
\toprule
\centering
Method & Precision & Latency (ms) \\
\midrule 
ChronoDepth & FP16 & 506 \\
DepthCrafter & FP16 & 910 \\
DepthAnyVideo & FP16 & 159 \\
NVDS & FP32 & 204 \\
DAv2-L & FP32 & 60 \\
VDA-L (Ours) & FP32 & 67 \\
VDA-S (Ours) & FP32 & \textbf{9.1} \\
\bottomrule
\end{tabular}
\caption{\textbf{Inference latency comparisons for video depth estimation.} We measure average runtime for each frame on a single A100 GPU with a resolution of $518 \times 518$.}
\label{tab::infer_time}
\vspace{-18pt}
\end{table}

\subsection{Ablation Studies}
Througout this section, we use the VDA-S model with a window size of 16, trained without image distillation unless otherwise specified. The metrics reported without a dataset name represent the mean values across all datasets.

\noindent{\bf{Temporal Loss.}} Temporal loss experiments are conducted on the TartanAir~\cite{wang2020tartanair} and VKITTI~\cite{cabon2020virtual} datasets. In addition to the TGM+SSI loss we proposed, we evaluate the performance of three other loss functions. The VideoAlign loss is a straightforward design that aligns predicted video depth to the ground truth using a shared scale-shift and computes the $l1$ loss. Building upon VideoAlign, the VideoAlign+SSI loss introduces an additional spatial loss to supervise the single-frame structure. The OPW+SSI loss combines optical flow-based warping loss proposed in~\cite{wang2023neural} with a single-frame spatial loss. SE refers to the stable error loss introduced in Equation~\ref{eq:stable_loss}. As shown in Table~\ref{tab::loss_ablation}, while VideoAlign and VideoAlign+SSI exhibit good geometric metrics, their video stability metrics are poor. Among loss functions with temporal constraints, our proposed TGM+SSI loss significantly outperforms the OPW+SSI loss on both geometric and stability metrics, and achieves metrics comparable to SE+SSI. It shows that TGM not only corrects the errors from OPW but also eliminates the dependency on optical flow.
\begin{table}[]
\centering
\resizebox{0.9\linewidth}{!}{
\begin{tabular}{lccc}
\toprule
\centering
Loss & AbsRel~(↓) & $\delta_1$~~(↑) & TAE~(↓) \\
\midrule 
VideoAlign & \textbf{0.151}  & \underline{0.846}  & 1.326\\
VideoAlign+SSI & \textbf{0.151}  & \textbf{0.848}  & 1.207 \\
OPW~\cite{wang2023neural}+SSI & 0.182  & 0.771  & 0.918  \\
SE+SSI & 0.160  & 0.836  & \textbf{0.753}  \\
TGM+SSI (Ours) & 0.166  & 0.832  & \underline{0.767}  \\
\bottomrule
\end{tabular}}
\caption{\textbf{Ablation studies on the effectiveness of the temporal losses.} 
``VideoAlign'' denotes the spatial loss with a shared scale-shift alignment applied to the entire video. ``SSI'' is the image-level spatial loss used in~\cite{depth_anything_v2}. ``OPW'' refers to the optical flow-based warping loss described in~\cite{wang2023neural}.
``SE'' refers to the stable error as introduced in Equation~\ref{eq:stable_loss}. ``TGM'' represents our proposed temporal gradient matching loss.
}
\label{tab::loss_ablation}
\vspace{-18pt}
\end{table}

\begin{figure}[t]
  \centering
   \includegraphics[width=1.0\linewidth]{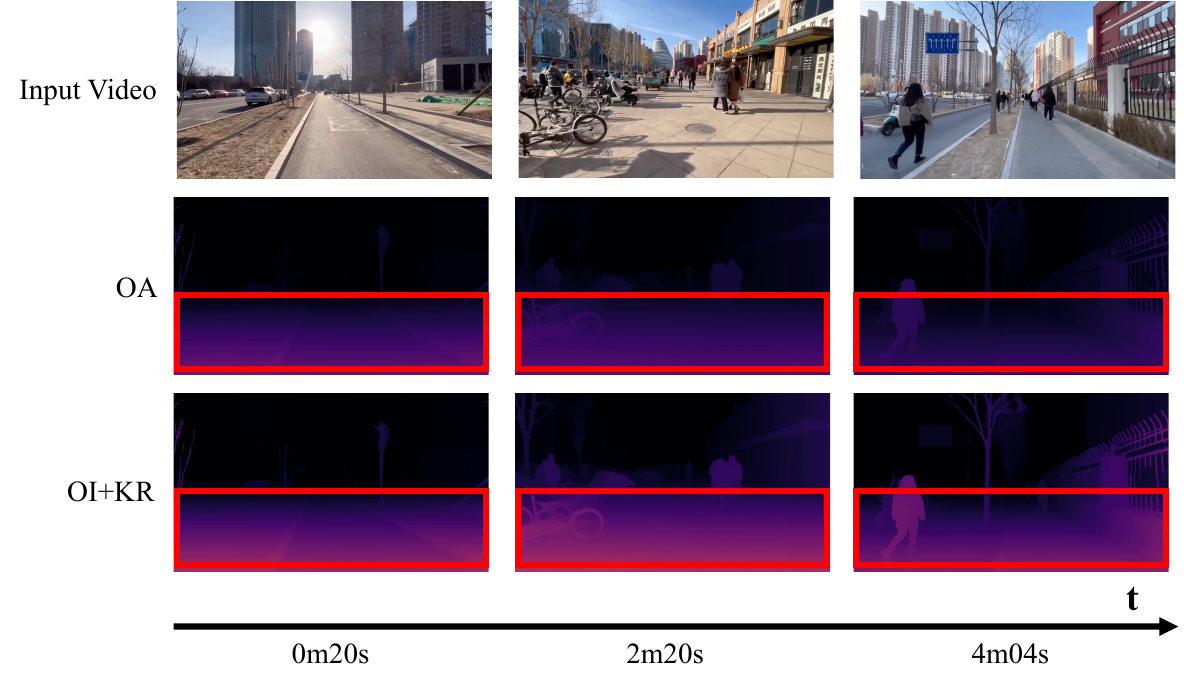}

   \caption{\textbf{Qualitative comparisons of different inference strategies.} We compare overlap alignment (OA) with our proposed overlap interpolation and key-frame referencing (OI + KR) on a self-captured video with 7320 frames.}
   \label{fig:infer_strategy_compare}
   \vspace{-8pt}
\end{figure}

\noindent{\bf{Inference Strategies.}} To ablate our inference strategy, we consider four different inference schemes. \textbf{Baseline}, inference is performed independently on each window without overlapping frames. \textbf{Overlap Alignment (OA)}, based on scale-shift invariant alignment of the overlapping frames between two neighboring windows, this method stitches the two windows together. \textbf{Overlap Interpolation (OI)}, following the approach in DepthCrafter~\cite{hu2024depthcrafter}, this method splices two windows together after performing linear interpolation in the overlap region. \textbf{Overlap Interpolation + KeyFrame Reference (OI+KR)}, on the basis of OI, we additionally introduce key frames from the previous window as a reference for the current inference. As shown in Table~\ref{tab::infer_strategy}, OA achieves metrics comparable to those of OI+KR. However, it leads to cumulative scale drift during long video inference. This issue is illustrated in Figure~\ref{fig:infer_strategy_compare}, where we evaluate OA and OI+KR on an extended video with a duration of $4^\prime 04^{\prime \prime}$. Notably, the red boxed region in the last frame of the video processed by OA highlights a cumulative drift in the depth scale. In contrast, OI+KR maintains global scale consistency more effectively throughout the duration of the video. One possible explanation for the better metrics of OA in the evaluation datasets is that the 500-frame evaluation video dataset is not long enough to reflect the scale drift issues encountered in real-world, long-duration videos.

\noindent{\bf{Window sizes.}} As shown in Table~\ref{tab::infer_strategy}, the model with a window size of 32 exhibits better geometric accuracy and temporal consistency compared to the model with a window size of 16. However, increasing the window size beyond 32 does not yield additional benefits. Given that a larger window size requires more resources for both training and inference, we select a window size of 32 for our final model.

\begin{table}[]
\centering
\begin{tabular}{lcccc}
\toprule
\centering
Strategy & Window & AbsRel~(↓) & $\delta_1$~~(↑) & TAE~(↓) \\
\midrule 
Baseline     & 16 & 0.157  & 0.826  & 0.874  \\
OA    & 16 & 0.146  & 0.845  & 0.792  \\
OI    & 16 & 0.157  & 0.826  & 0.783  \\
OI+KR(Ours) & 16 & 0.145  & 0.849  & 0.761  \\
OI+KR(Ours) & 32 & \underline{0.144}  & \underline{0.851}  & \textbf{0.718} \\
OI+KR(Ours) & 48 & \textbf{0.143}  & \textbf{0.852}  & \underline{0.732} \\
\bottomrule
\end{tabular}
\caption{\textbf{Ablation studies on the effectiveness of different inference strategies and window sizes.} ``Baseline'' denotes directly inference for video clips without overlapping. ``OA'' denotes inference with a overlap of 4 frames and perform scale-shift alignment across windows. ``OI'' denotes depth clip stitching with a overlap of 4 frames. ``OI+KR'' combines the ``OI'' with our proposed key-frame referencing with extra 2 key-frames.}
\label{tab::infer_strategy}
\vspace{-8pt}
\end{table}

\noindent{\bf{Training Strategies.}} In addition to training on synthetic datasets, we conduct an ablation study of distillation training strategies by incorporating an equal amount of pseudo-labeled real datasets. As shown in Table~\ref{tab::train_strategy_ablation}, the inclusion of real single-frame datasets in the distillation training process results in a notable enhancement of single-frame depth metrics in both AbsRel and $\delta_1$. Furthermore, it also improves video depth metrics.
\begin{table}[]
\centering
\resizebox{0.9\linewidth}{!}{
\begin{tabular}{lccccc}
\toprule
\centering
\multirow{2}{*}{Datasets} & \multicolumn{2}{c}{Image-datasets} & \multicolumn{3}{c}{Video-datasets} \\ 
\cmidrule(r){2-6} 
& AbsRel~(↓) & $\delta_1$~(↑) & AbsRel~(↓) & $\delta_1$~(↑) & TAE~(↓) \\
\midrule 
Video & 0.180 & 0.876 & 0.145 & 0.849 & 0.761  \\
Video + Image & \textbf{0.167} & \textbf{0.883} & \textbf{0.142} & \textbf{0.852} & \textbf{0.742}  \\
\bottomrule
\end{tabular}}
\caption{\textbf{Ablation studies on the effectiveness of the image dataset distillation.} ``Video'' denotes training using only video datasets. ``Video + Image'' merges video and image datasets for training using image-level distillation~\cite{depth_anything_v2}.}
\label{tab::train_strategy_ablation}
\vspace{-18pt}
\end{table}
\section{Conclusion}
\label{sec:conclusion}

In this paper, we present a novel method named \name for estimating the depth of the video that is temporally consistent. The model is built on top of Depth Anything V2 and is based on three key components. First, a spatial-temporal head to involve temporal interactions by applying a temporal self-attention layer to feature maps. Second, a simple temporal gradient matching loss function is used to enforce temporal consistency. Third, to enable long-video depth estimation, a novel keyframe-based strategy is developed for segment-wise inference along with a depth stitching method. Extensive experiments show that our model achieves state-of-the-art performance in three aspects: spatial accuracy, temporal consistency, and computational efficiency. Consequently, it can produce high-quality depth predictions for videos lasting several minutes.


{\small
\bibliographystyle{ieee_fullname}
\bibliography{egbib}
}
\clearpage
\maketitlesupplementary
\setcounter{section}{0}

\section{More Qualitative Results}

We present more qualitative comparisons among different approaches for static images and evaluation videos.

\begin{figure}[bht]
    \includegraphics[width=\linewidth]{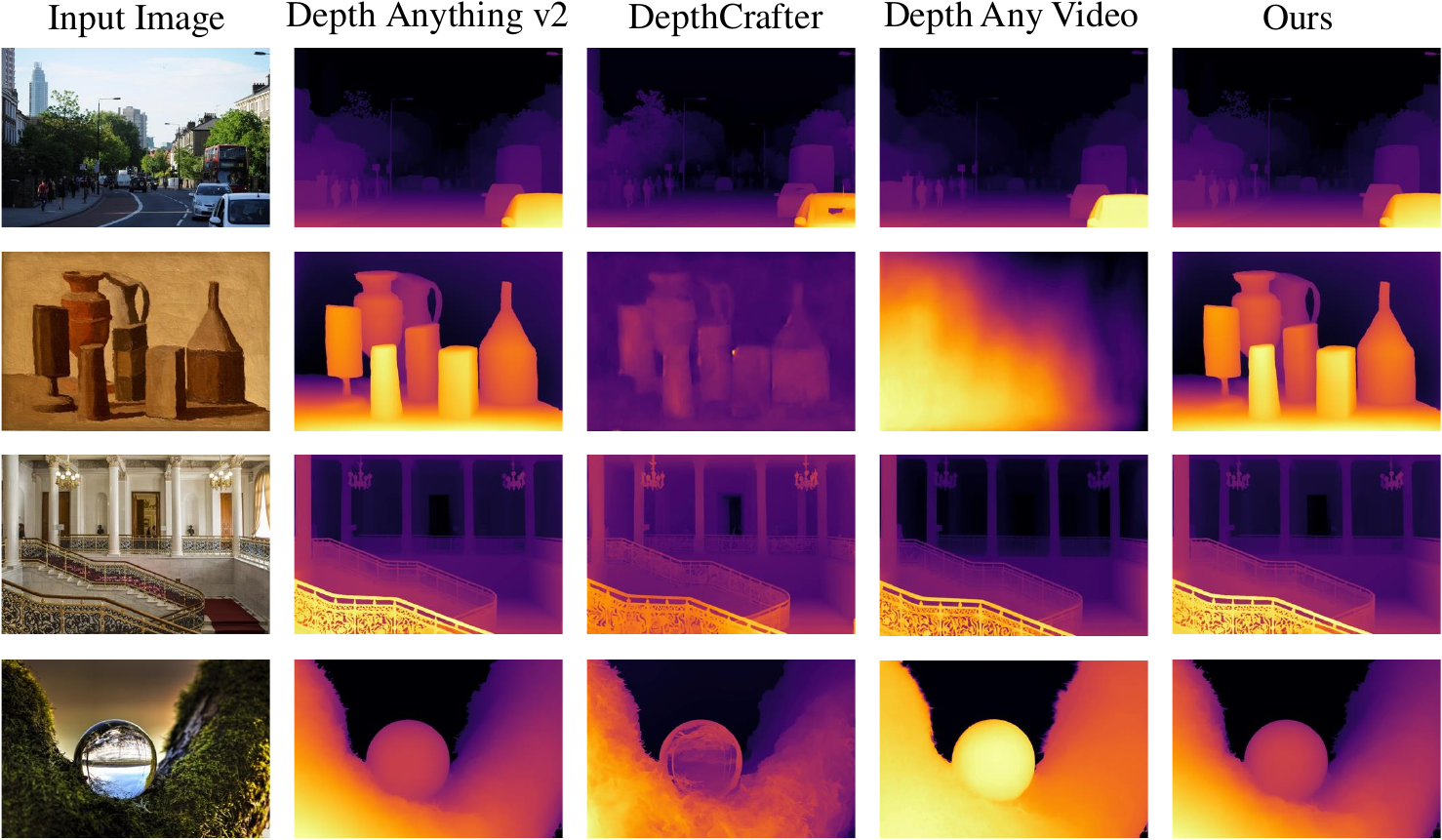}
    \captionof{figure}{\textbf{Qualitative comparison for static image depth estimation.} We compare our model with Depth-Anything-V2~\cite{depth_anything_v2}, DepthCrafter~\cite{hu2024depthcrafter}, and Depth Any Video~\cite{depthanyvideo} on static image depth estimation. Our model demonstrates visualization results comparable to those of Depth-Anything-V2~\cite{depth_anything_v2}.}
    \label{fig:single_frame_show_sup}
    \vspace{-18pt}
\end{figure}

\paragraph{In-the-wild image results.} Static image depth estimation results are shown in Fig. \ref{fig:single_frame_show_sup}. DepthCrafter~\cite{hu2024depthcrafter} and Depth Any Video~\cite{depthanyvideo} exhibit poor performance on oil paintings. DepthCrafter~\cite{hu2024depthcrafter} also struggles with transparent objects such as glass and water. Compared with these methods, our model demonstrates superior depth estimation results in complex scenarios. Moreover, our model shows depth estimation results for static images that are comparable to those of Depth-Anything-V2~\cite{depth_anything_v2}, demonstrating that we have successfully transformed Depth-Anything-V2 into a video depth model without compromising its spatial accuracy.

\begin{figure*}
    \centering
    \includegraphics[width=\linewidth]{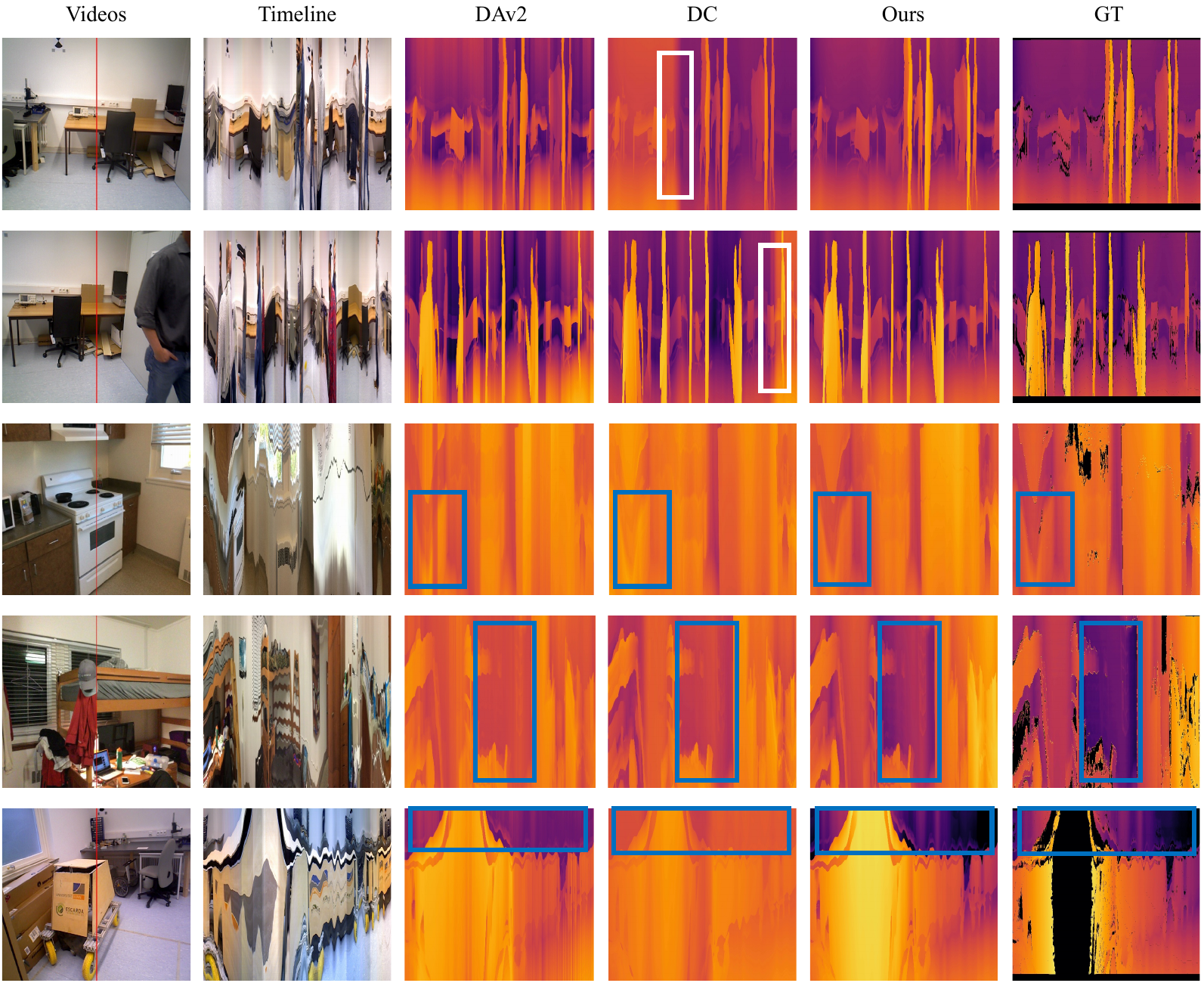}
    \caption{\textbf{Qualitative comparison for real-world long video depth estimation.} We compare with Depth-Anything-V2~\cite{depth_anything_v2} and DepthCrafter~\cite{hu2024depthcrafter} on 500-frames videos from Scannet~\cite{dai2017scannet} and Bonn~\cite{palazzolo2019iros} . We show changes in color and depth over time at the vertical red line in videos. White boxes show inconsistent estimation. Blue boxes show our algorithm has higher accuracy. }
    \label{fig:long_show_sup}
    \vspace{-8pt}
\end{figure*}

\paragraph{Evaluation video results.} We showcase five video visualization results from the evaluation datasets Scannet~\cite{dai2017scannet} and Bonn~\cite{palazzolo2019iros} in Fig. \ref{fig:long_show_sup}. For enhanced visualization, all predicted video depths are aligned to the ground truth video depths using the same method as in the evaluation. DepthCrafter~\cite{hu2024depthcrafter} exhibits depth drift in long videos, as indicated by the blue boxes. Moreover, our model demonstrates superior depth accuracy compared to DepthCrafter~\cite{hu2024depthcrafter}, as highlighted in the red boxes.

\section{Short video depth quantitative results} We compare our model with DepthCrafter~\cite{hu2024depthcrafter} and Depth Any Video~\cite{depthanyvideo} on the KITTI~\cite{geiger2013vision}, Bonn~\cite{palazzolo2019iros}, and Scannet~\cite{dai2017scannet} datasets, with frame lengths of 110, 110, and 90, respectively, corresponding to the settings in~\cite{hu2024depthcrafter}. As shown in Tab.~\ref{tab::short_video_benchmark}, our model demonstrates a significant advantage of approximately 7\% over both DepthCrafter~\cite{hu2024depthcrafter} and Depth Any Video~\cite{depthanyvideo} on the Scannet dataset~\cite{dai2017scannet}. On the KITTI dataset~\cite{geiger2013vision}, our model significantly outperforms DepthCrafter~\cite{hu2024depthcrafter} by about 7\%. Additionally, our model achieves comparable results on Bonn~\cite{palazzolo2019iros} and KITTI~\cite{geiger2013vision} compared to Depth Any Video~\cite{depthanyvideo}. It is worth noting that the parameters of our model and the video depth data used for training are significantly smaller than those of DepthCrafter~\cite{hu2024depthcrafter} and Depth Any Video~\cite{depthanyvideo}, demonstrating the effectiveness and efficiency of our method.

\begin{table*}[]
\centering
\resizebox{0.9\linewidth}{!}{
\begin{tabular}{lcccccccc}
\toprule
\centering
\multirow{2}{*}{Method / Metrics} & \multirow{2}{*}{Params(M)} & \multirow{2}{*}{\# Video Training Data(M)} & \multicolumn{2}{c}{KITTI(110)~\cite{geiger2013vision}} & \multicolumn{2}{c}{Bonn(110)~\cite{palazzolo2019iros}} & \multicolumn{2}{c}{Scannet(90)~\cite{dai2017scannet}} \\ 
\cmidrule(r){4-9} 
& & & AbsRel~(↓) & $\delta_1$~(↑) & AbsRel~(↓) & $\delta_1$~~(↑) & AbsRel~(↓) & $\delta_1$~~(↑) \\
\midrule 
DepthCrafter & 2156.7 & 10.5\raisebox{0.5ex}{\texttildelow}40.5 & 0.111  & 0.885  & 0.066  & \underline{0.979}  & 0.125  & 0.848 \\
DepthAnyVideo & 1422.8 & 6 & \textbf{0.073}  & \textbf{0.957}  & \textbf{0.051}  & \textbf{0.981}  & \underline{0.112}  & \underline{0.883} \\
VDA-L (Ours) & 381.8 & 0.55 & \underline{0.079}  & \underline{0.950}  & \underline{0.053}  & 0.972  & \textbf{0.075}  & \textbf{0.954} \\
\bottomrule
\end{tabular}}
\caption{\textbf{Zero-shot short video depth estimation results}. We compare with DepthCrafter~\cite{hu2024depthcrafter} and DepthAnyVideo~\cite{depthanyvideo} in short video depth benchmark. ``VDA-L'' denotes our model with ViT-Large backbone. The default inference resolution of our model is set to 518 pixels on the short side, maintaining the aspect ratio. The \textbf{best} and the \underline{second best} results are highlighted.}
\label{tab::short_video_benchmark}
\end{table*}

\section{Limitations and future work}
Our model is trained primarily on publicly available video depth datasets, which may limit its capabilities due to the data quantity. We believe that with more data, the model's performance can be further improved, and the backbone network can be unlocked for fine-tuning. Additionally, although our model is significantly more computationally efficient than the baselines, it still faces challenges in handling streaming videos, which we leave as future work.

\section{More Details of Pipeline}
\paragraph{Spatiotemporal head details.} 

Among the four temporal layers, two are inserted after the Reassemble layers at the two smallest resolutions, and the other two are inserted before the last two Fusion layers.

\begin{figure}
    \centering
    \includegraphics[width=0.4\linewidth]{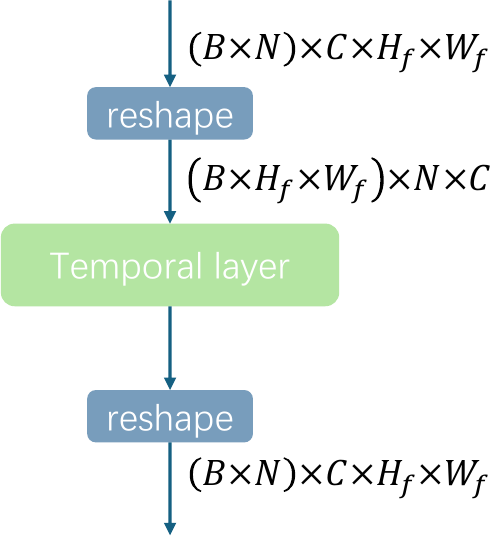}
    \caption{\textbf{Temporal layer.} The feature shape is adjusted for temporal attention.}
    \label{fig:head_detail}
    \vspace{-18pt}
\end{figure}

The shape of the feature is transformed into $(B \times H_f \times W_f) \times N \times C$ before each temporal layer and is transformed back to $(B \times N) \times C \times H_f \times W_f$ after each temporal layer. Here, $B$ denotes the batch size, $N$ represents the number of frames in the video clip, $H_f$ and $W_f$ are the height and width of the feature, respectively, and $C$ represents the number of channels in the feature, as shown in Fig. \ref{fig:head_detail}

\paragraph{Image distillation details.} We follow the approach in~\cite{depth_anything_v2} and use a teacher model that comprises a ViT-giant encoder and is trained on synthetic datasets. The loss function used for distillation is identical to the spatial loss employed for video depth data.

\paragraph{Training dataset details.} For video training, we utilize four synthetic datasets with precise depth annotations: TartanAir~\cite{wang2020tartanair}, VKITTI~\cite{cabon2020virtual}, PointOdyssey~\cite{zheng2023pointodyssey}, and IRS~\cite{wang2021irslargenaturalisticindoor}, totally 0.55 million frames. The TartanAir~\cite{wang2020tartanair}, VKITTI~\cite{cabon2020virtual}, PointOdyssey~\cite{zheng2023pointodyssey}, and IRS~\cite{wang2021irslargenaturalisticindoor} datasets contain 0.31M, 0.04M, 0.1M, and 0.1M frames, respectively. Additionally, 0.18 million frames from wild binocular videos labeled with~\cite{jing2024match-stereo-videos} are included for training. We also incorporate a subset of real-world unlabeled datasets from~\cite{depth_anything_v2} for single image supervision, totaling 0.62 million frames. Notably, we excluded 0.13M frames from PointOdyssey~\cite{zheng2023pointodyssey} that do not contain background depth ground truth, resulting in our usage of only half of the original dataset. Due to the uneven data distribution across the four training datasets, we employ a uniform sampler to ensure that each dataset contributes equally during training.

\paragraph{Implementation Details}
The weights are initialized from Depth Anything V2~\cite{depth_anything_v2}. Training comprises two stages. In the first stage, synthetic and wild binocular data are used. In the second stage, synthetic videos and unlabeled single images are employed. Models labeled '-Syn' in the main paper are exceptionally trained using synthetic videos and unlabeled images in a single stage. Besides the loss defined in Equation 4 of the main paper used for synthetic videos, unlabeled single images are supervised using the same method as described in~\cite{depth_anything_v2}.
During training, we uniformly sample video clips of 32 frames from each dataset, resize the shorter edge of images to 518 pixels, and perform random center cropping, resulting in training clips with a resolution of $518 \times 518 \times 32$. We use the AdamW~\cite{loshchilov2017decoupled} optimizer with a cosine scheduler, setting the base learning rate to $1e^{-4}$. The batch size is set to 16 for video frames, each with a length of 32 frames, and 128 for image datasets. The loss weights for the single frame loss, TGM loss, and distillation loss are set to 1.0, 10.0, and 0.5, respectively.
\label{impl_detail}

\section{More Details of Evaluation}

\paragraph{Evaluation dataset details.} We use a total of five datasets for video depth evaluation: KITTI~\cite{geiger2013vision}, Scannet~\cite{dai2017scannet}, Bonn~\cite{palazzolo2019iros}, NYUv2~\cite{Silberman:ECCV12}, and Sintel~\cite{Butler_Wulff_Stanley_Black_2012}. Specifically, we use Scannet~\cite{dai2017scannet} and NYUv2~\cite{Silberman:ECCV12} for static indoor scenes, Bonn~\cite{palazzolo2019iros} for dynamic indoor scenes, KITTI~\cite{geiger2013vision} for outdoor scenes, and Sintel~\cite{Butler_Wulff_Stanley_Black_2012} for wild scenes. For NYUv2~\cite{Silberman:ECCV12}, we sample 8 videos from the original dataset, which contains 36 videos. Our evaluation comprises three different settings: long videos, long videos with different frame lengths, and short videos. For the long video evaluation, we use all five datasets and set the maximum frame length to 500 for each video. For the evaluation of long videos with different frame lengths, we select subsets of videos with frame lengths greater than 500 from Scannet~\cite{dai2017scannet}, Bonn~\cite{palazzolo2019iros}, and NYUv2~\cite{Silberman:ECCV12}. For the short video evaluation, we use KITTI~\cite{geiger2013vision}, Bonn~\cite{palazzolo2019iros}, and Scannet~\cite{dai2017scannet}, setting the maximum frame lengths to 110, 110, and 90, respectively, in accordance with the settings in DepthCrafter~\cite{hu2024depthcrafter}. In addition to video depth evaluation, we also assess our model's performance on static images. Following ~\cite{depth_anything_v2}, we perform evaluations on five image benchmarks: KITTI~\cite{geiger2013vision}, Sintel~\cite{Butler_Wulff_Stanley_Black_2012}, NYUv2~\cite{Silberman:ECCV12}, ETH3D~\cite{Schops_Schonberger_Galliani_Sattler_Schindler_Pollefeys_Geiger_2017}, and DIODE~\cite{Igor_Nicholas_Zhang_Luo_Wang_Dai_Daniele_Mostajabi_Basart_Walter_et_al_2019}. To ensure a fair comparison, all evaluation videos and images are excluded from the training datasets.

\paragraph{Evaluation metric details.} All video metrics we evaluated are based on ground truth depth. Specifically, we use the least squares method to compute the optimal scale and shift to align the entire inferred video inverse depth with the ground truth inverse depth. The aligned inferred video inverse depth is then transformed into depth, which is subsequently used to compute the video metrics with the ground truth depth. For geometric accuracy, we compute the Absolute Relative Error (AbsRel) and $\delta_1$ metrics, following the procedures outlined in ~\cite{depth_anything_v2,hu2024depthcrafter}. To assess temporal stability, we use the Temporal Alignment Error (TAE) metric in ~\cite{depthanyvideo}, to measure the reprojection error of the depth maps between consecutive frames. We use Equation \ref{eq:correct_tae}.
\begin{equation}
\begin{aligned}
TAE=  \frac {1}{2(N-1)}  &\sum _ {k=1}^ {N-1}  AbsRel(f(  \hat {x}_ {d}^ {k}  ,  p^ {k}  ),  \hat {x}_ {d}^ {k+1}  ) + \\
&AbsRel(f(  \hat {x}_ {d}^ {k+1}  ,  p^ {k+1}_-  ),  \hat {x}_ {d}^ {k}  )
\end{aligned}
\label{eq:correct_tae}
\end{equation}

Here, $f$ represents the projection function that maps the depth $\hat{x}^k_d$ from the $k$-th frame to the $(k+1)$-th frame using the transformation matrix $p^k$. $p^{k+1}_-$ is the inverse matrix for inverse projection. $N$ denotes the number of frames.

\paragraph{Baseline implementations. } We obtain the inferences of DepthCrafter~\cite{hu2024depthcrafter}, Depth Any Video~\cite{depthanyvideo}, and NVDS~\cite{wang2023neural} using the respective inference code provided by the authors. Specifically, DepthCrafter~\cite{hu2024depthcrafter} employs different inference resolutions for different datasets. Depth Any Video~\cite{depthanyvideo} infers with a maximum dimension of 1024. NVDS~\cite{wang2023neural} performs inference on a video twice, with a minimum dimension of 384, once in the forward direction and once in the backward direction, and computes the mean result from these two passes. For Depth-Anything-V2~\cite{depth_anything_v2}, we obtain the video depth results by inferring each frame individually with a minimum dimension of 518.

\section{Applications}
\begin{figure*}[t]
   \includegraphics[width=1\linewidth]{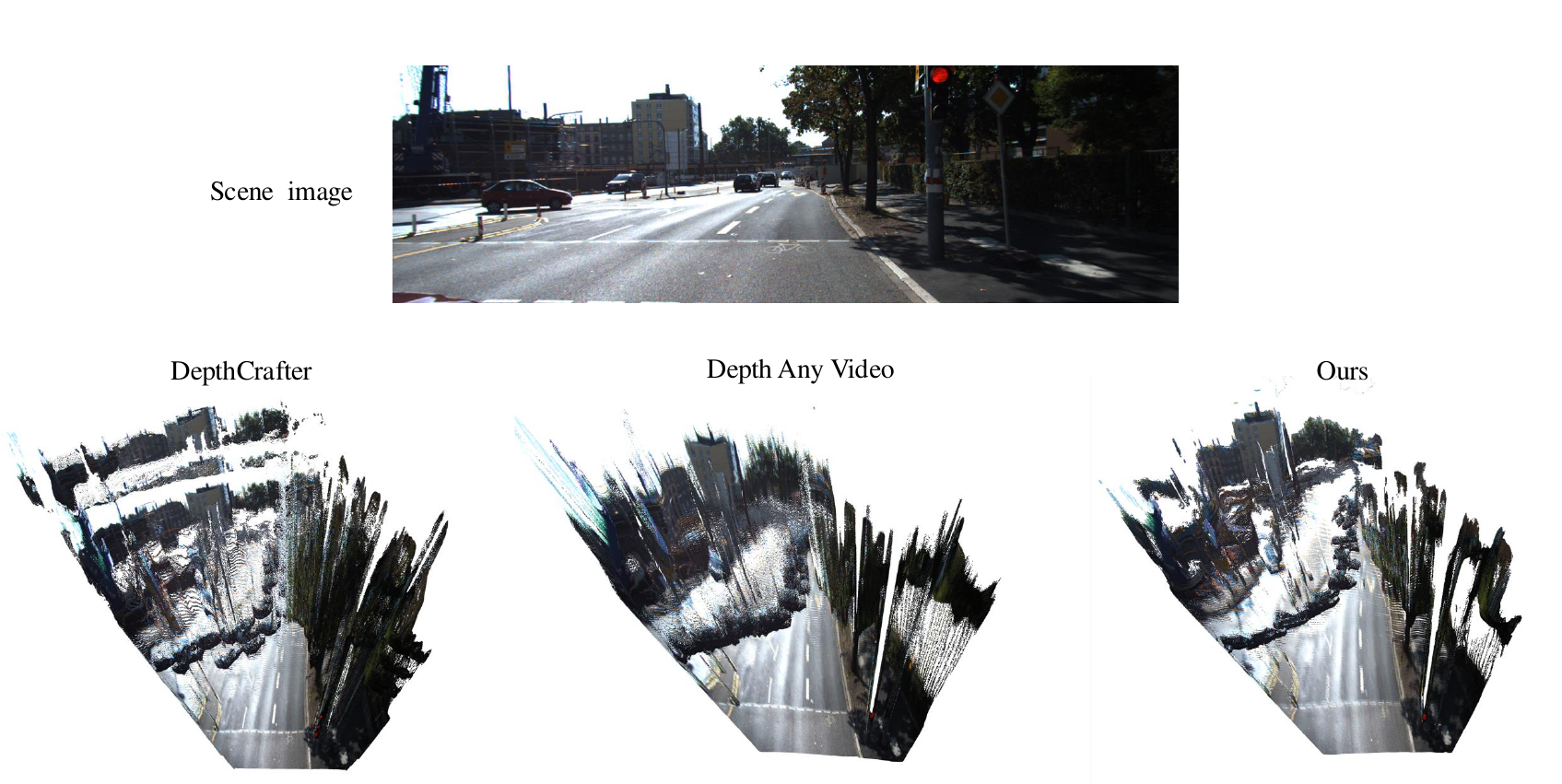}

   \caption{\textbf{Dense point cloud generation.} We compare our model with DepthCrafter~\cite{hu2024depthcrafter} and DepthAnyVideo~\cite{depthanyvideo} for dense point cloud generation on the KITTI dataset~\cite{geiger2013vision}. Our model generates a clean and regular point cloud from multiple frames spanning approximately 5 seconds. In contrast, the point cloud generated by DepthCrafter~\cite{hu2024depthcrafter} contains several obvious discontinuous layers. DepthAnyVideo~\cite{depthanyvideo} produces a point cloud with numerous noisy outliers and noticeable distortion in distant views.}
   \label{fig:application_pcd}
   \vspace{-18pt}
\end{figure*}

\begin{figure*}[t]
   \includegraphics[width=0.85\linewidth]{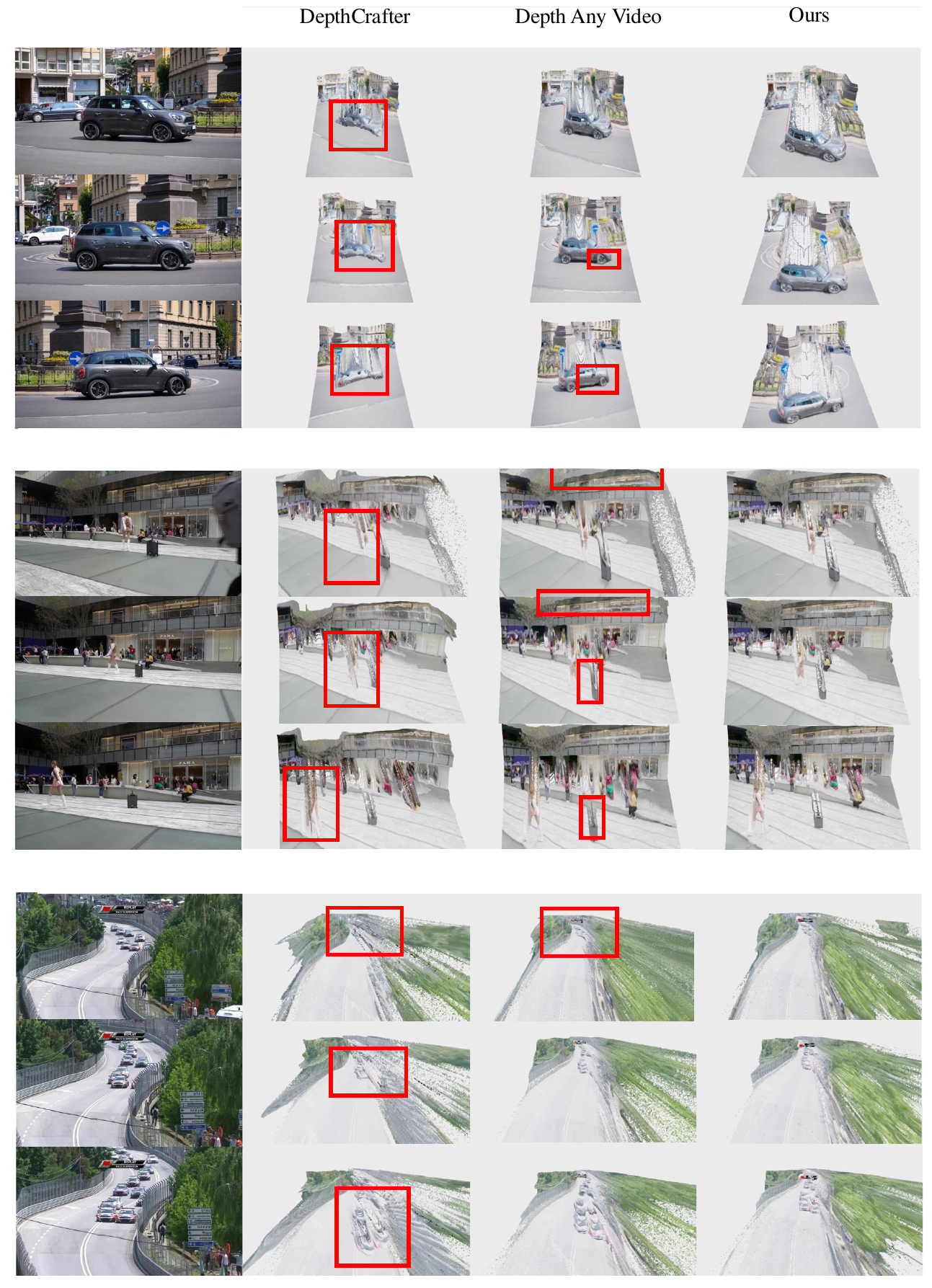}

   \caption{\textbf{Point cloud generation for wild videos.} 
   We compare our method with DepthCrafter~\cite{hu2024depthcrafter} and DepthAnyVideo~\cite{depthanyvideo} using three videos from DAVIS dataset~\cite{Perazzi2016davis}. Camera intrinsics, along with aligned scale and shift parameters, are derived from processing the first frame of each video through MoGe~\cite{wang2024moge}. Point cloud distortions are highlighted with red boxes.
}
   \label{fig:application_pcd_wild}
\end{figure*}

\paragraph{Dense point cloud generation.} By aligning single frame with metric depth, which can be obtained from a metric depth model or a sparse point cloud acquired through SLAM, our model can generate a depth point cloud for the entire environment using camera information. The generated point cloud can then be transformed into a mesh and utilized for 3D reconstruction, AR, and VR applications. We present a point cloud generation case in Fig. \ref{fig:application_pcd}. Here, we sample 10 frames spanning approximately 5 seconds from the KITTI dataset~\cite{geiger2013vision}. After obtaining the inferred inverse depth, we compute the global scale and shift by aligning the first frame with the corresponding metric inverse depth. We then apply the affine transformation to the entire set of inverse depth frames and convert them to depth. The final point cloud is generated by merging the point clouds from each frame. As shown in Fig. \ref{fig:application_pcd}, our model generates a clean and regular point cloud compared to DepthCrafter~\cite{hu2024depthcrafter} and Depth Any Video~\cite{depthanyvideo}. Point cloud generation for wild videos is illustrated in Fig. \ref{fig:application_pcd_wild}. Compared to DepthCrafter~\cite{hu2024depthcrafter} and DepthAnyVideo~\cite{depthanyvideo}, our model produces more regular point clouds.

\begin{figure}[t]
  \centering
   \includegraphics[width=1\linewidth]{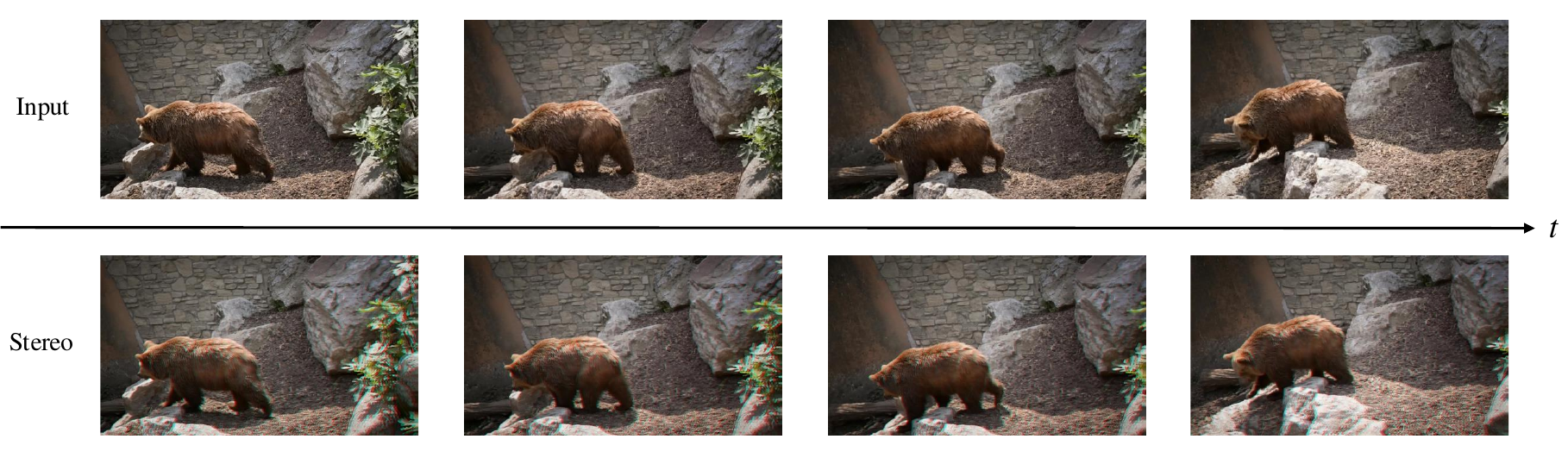}

   \caption{\textbf{3D Video Conversion.} A video from the DAVIS dataset~\cite{Perazzi2016davis} is transformed into a 3D video using our model.}
       \label{fig:app_stereo}
    \vspace{-18pt}
\end{figure}

\paragraph{3D Video Conversion.} Our model can be used to generate 3D videos. Compared to 3D videos generated by monocular depth models, those produced by our video depth model exhibit smoother and more consistent 3D effects. An example is presented in Fig.\ref{fig:app_stereo}.

\end{document}